# SPARSE POINT-GUIDED FUSION OF SUPERVISED AND SELF-SUPERVISED LEARNING MODEL FOR SEAWEED SEGMENTATION

**Tatsuya Suzuki**
Fujitsu Limited
Kawasaki, Japan

**Kazuya Ijuin**
Fujitsu Limited
Kawasaki, Japan

**Hideki Tomimori**
Fujitsu Limited
Kawasaki, Japan

**Megumi Chikano**
Fujitsu Limited
Kawasaki, Japan

**Katsushi Sakai**
Fujitsu Limited
Kawasaki, Japan

## ABSTRACT

The ocean plays a critical role in sustainable development, particularly in climate change mitigation. Among marine ecosystems, blue carbon ecosystems are recognized as important natural carbon sinks. In this context, this paper addresses precise seaweed classification for blue carbon quantification in Ocean Digital Twin initiatives. Conventional methods, including supervised learning (limited by data scarcity and domain gaps) and self-supervised learning (unable to assign class labels), struggle with underwater complexities and diverse seaweed species. To overcome this, we propose a novel two-stage seaweed segmentation technique. This technique first utilizes Supervised and Self-supervised Learning Model Propagation (SSL.Prop.), which leverages supervised learning for initial class information and approximate locations, guiding self-supervised learning for detailed, accurate segmentation. Subsequently, MaskFusion (MF) refines these results by merging instance-level masks for highly accurate segmentation. This integrated approach allows automatic class label assignment and mitigates domain gap effects. Specifically, instance segmentation estimates sparse point locations which then guide self-supervised learning for detailed region segmentation. Evaluated with underwater images from Yamaguchi Prefecture, our full proposed method (SSL.Prop.+MF) achieved a 0.082 mIoU improvement over USIS-SAM, demonstrating significant accuracy gains, particularly for small seaweed. This approach demonstrates strong potential for improving blue carbon quantification and marine ecosystem monitoring.

## 1. INTRODUCTION

In recent years, the multifaceted utilization of the ocean has attracted attention as a novel solution to global societal challenges such as climate change and sustainable resource use. The ocean, serving as a foundation for diverse fields including energy, environment, ecosystems and logistics, is a critical domain for realizing the sustainable development of humankind. To maximize its potential, it is essential to accurately grasp the state of the ocean and quantitatively understand its changes. This enables the formulation and implementation of optimal measures tailored to specific objectives. Against this backdrop, we aim to realize the "Ocean Digital Twin [1]" that integrates data collection, analysis and policy deliberation for all aspects of the ocean (FIGURE 1). This is an innovative approach that reproduces the real ocean in a virtual space, supporting decision-making through various simulations and predictions.

As a first step towards realizing the Ocean Digital Twin, this research focuses on the utilization of blue carbon credits from seaweed beds in coastal areas of Japan. Seaweed beds are globally recognized for their importance as blue carbon, absorbing and fixing $CO_2$ which contributes to global warming. Accurate quantification of blue carbon ($CO_2$ absorption and fixation) is extremely important for assessing the health of seaweed beds, visualizing carbon absorption and calculating the economic value of marine ecosystem services. To enable this assessment, techniques for highly accurate and automatic identification of seaweed distribution, species and coverage from underwater images are indispensable. However, factors specific to underwater environments, such as turbidity, rapid light attenuation, and the diversity of seaweed, especially the presence of small, difficult-to-distinguish species, present challenges.

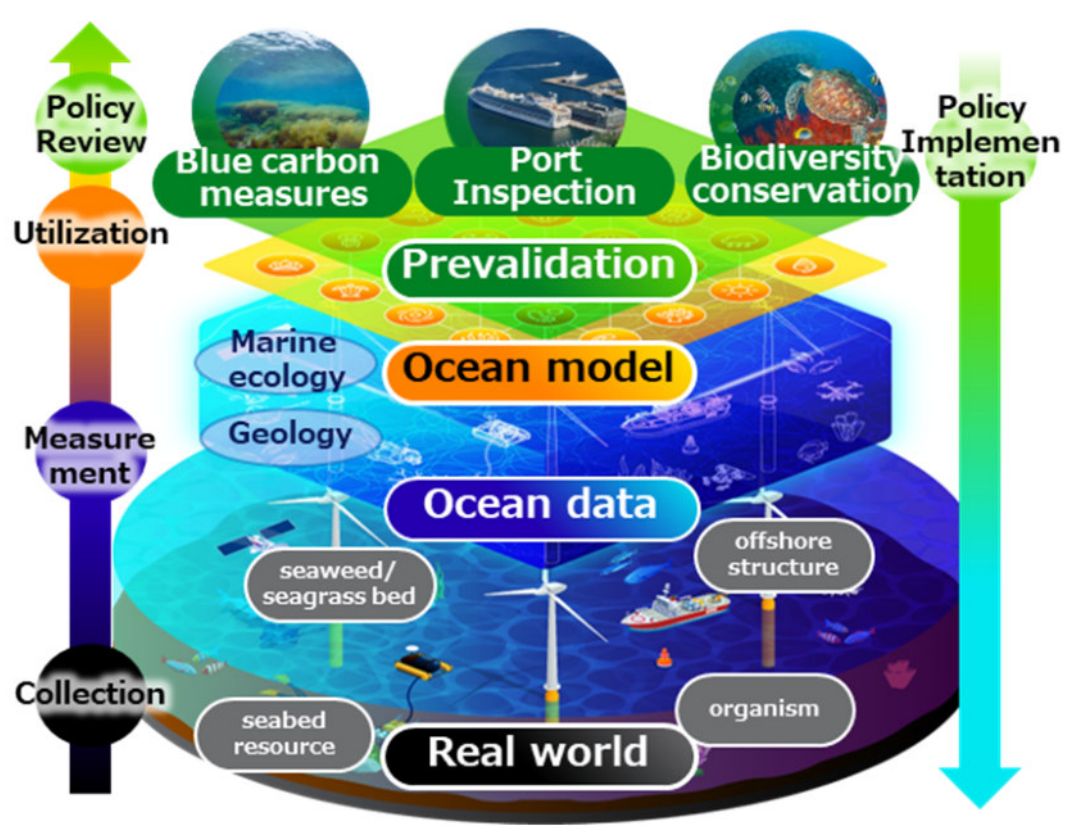

**FIGURE 1:** OCEAN DIGITAL TWIN

## 2. RELATED WORK

Automated seaweed classification techniques are actively researched in the field of computer vision, particularly for semantic and instance segmentation tasks that classify objects pixel-wise in images. Existing segmentation techniques primarily fall into two categories: supervised learning, which classifies test images from learned images, and self-supervised learning, which compares local regions within test images (FIGURE 2). These are described in detail below.

**Supervised Learning:** This method involves training an AI using ground truth data (images and labels indicating what each pixel represents). Examples of models include SUIM-Net [2] for semantic segmentation and USIS-SAM [3] for instance segmentation. This approach excels at classifying images similar to those used for training. However, the performance of supervised learning heavily depends on the quality and quantity of available ground truth data. Particularly, when the training data contains a limited number of images or variations of specific seaweed species, the classification accuracy for those species significantly decreases.

**Self-Supervised Learning:** This approach involves generating training signals from large-scale unlabeled image data for the model to learn. It utilizes internal features of the image (color, texture, shape, etc.) to segment similar regions, allowing for robust segmentation in unknown environments and diverse scenes. Examples of models include DINOv2 [4], Denoising-ViT [5], and HIL-Coral [6] as a seaweed classification technique utilizing these models. A major advantage of this approach is its significant reduction in annotation costs. However, self-supervised learning primarily generates region segmentation based on visual features within the image, making it difficult to automatically assign specific class labels, such as identifying the species of seaweed within each segmented region. Therefore, even if automatic region segmentation is possible, it does not directly lead to accurate quantitative assessment for blue carbon evaluation, resulting in a challenge that the final process cannot be fully automated.

Thus, existing techniques face the challenge of achieving both high classification accuracy and complete automation.

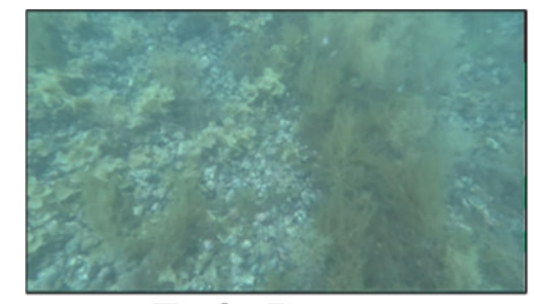

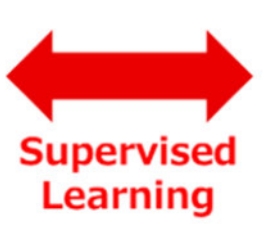

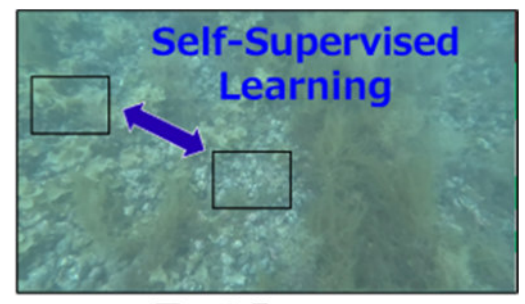

**FIGURE 2:** SUPERVISED LEARNING AND SELF-SUPERVISED LEARNING

## 3. PROPOSED METHODS

The novel seaweed segmentation technique proposed in this research not only combines supervised and self-supervised learning models but also intelligently integrates and complements their outputs through a two-stage fusion strategy to maximize the benefits of each. FIGURE 3 illustrates the overall scheme of the proposed method. This strategic combination is designed to overcome the inherent limitations of each individual approach, ensuring both high accuracy and automation. The subsequent sections detail the individual components and their synergistic integration.

### 3.1 Supervised Learning Model

In the first stage, we utilize a supervised learning model to perform instance segmentation of seaweed from underwater images. Specifically, for an input underwater image, we generate BBoxes (Bounding Boxes) indicating where each seaweed species exists and masks (filled regions) representing its specific shape. This process provides approximate spatial and semantic information about individual seaweed instances and their species (e.g., this region contains *Sargassum piluliferum*). In this research, we use the USIS-SAM model, which is trained specifically for this task. USIS-SAM can perform both object detection and segmentation simultaneously, providing efficient initial recognition and a basis for the sparse point information required in later steps. This initial supervised step is crucial for providing strong semantic guidance, which is then refined by the self-supervised component, especially in complex underwater scenes.

### 3.2 Self-Supervised Learning Model

Next, we aim to generate more detailed segmentation masks using a self-supervised learning model-based technique than the information obtained in the previous stage. In this stage, in addition to the input underwater image, sparse point labels representing feature points within the image are used. The self-supervised learning model captures finer features within the image based on this information and then creates more precise segmentation masks through label propagation. In this research, we use DINOv2, a powerful self-supervised learning model, as a deep feature extraction backbone and combine it with HIL-Coral, a label propagation technique that utilizes these features. DINOv2, pre-trained on the large-scale LVD-142M dataset [4], combined with HIL-Coral for efficient label propagation, enables detailed, precise segmentation. This powerful combination allows for robust analysis of local image characteristics, leading to highly granular segmentation that is adaptable to diverse underwater conditions.

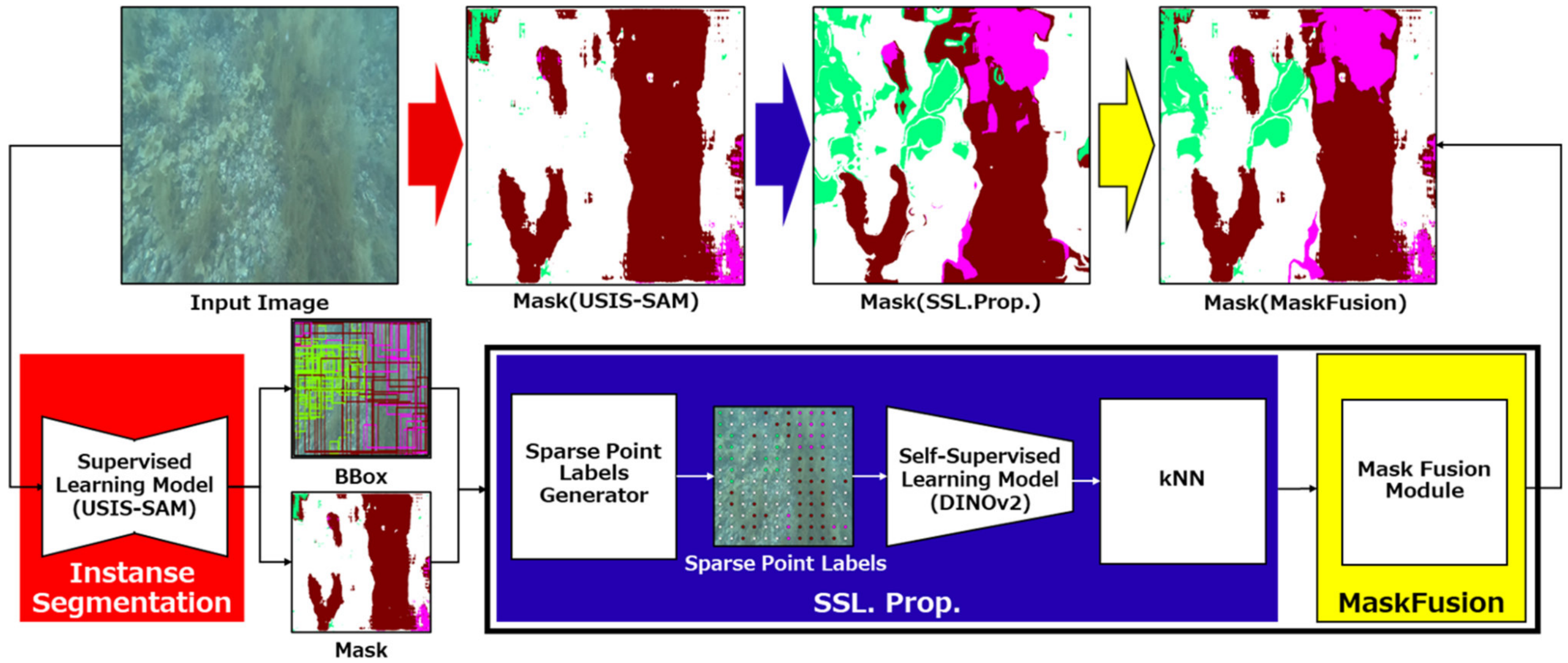

**FIGURE 3:** PROPOSED METHOD OVERVIEW

### 3.3 Fusion Strategy

The core of the proposed method involves effectively fusing the two aforementioned models to create a robust segmentation framework for challenging underwater imagery. This research proposes two distinct fusion methods: SSL.Prop. and MaskFusion. Prior to detailing each method, this section outlines the common fusion strategy.

#### 3.3.1 SSL. Prop.

Supervised and Self-supervised Learning Model Propagation (SSL.Prop.) links initial recognition by a supervised learning model with detailed segmentation by a self-supervised learning model. Beginning with BBox and mask information from the supervised learning model (USIS-SAM) for initial detection and approximate segmentation, the overall procedure within SSL.Prop. is summarized as follows (FIGURE 4)

**STEP 1. Small Algae BBox Filtering:** First, instances are prioritized by filtering BBoxes from the supervised model. We target BBoxes with a confidence score $s(B)$ above $T_s$ (e.g., $T_s$ = 0.8) and an area $A(B)$ less than $R_{small}$ (e.g., $R_{small}$ = 0.1) relative to the total image area $A_{img}$ . Only BBoxes satisfying these criteria are considered for sparse point generation, focusing on challenging instances like small seaweed. This selection criterion is:

$$B_p = \{B | s(B) \geq T_s \wedge A(B)/A_{img} \leq R_{small}\} \quad (1)$$

**STEP 2. Sparse Point Generation:** For each filtered BBox, Sparse Points are placed in a $G \times G$ grid. We use $G$ =12, generating 144 Sparse Point Labels per BBox (see Section 4.2).

**STEP 3. Label Refinement by BBox:** Subsequently, generated Sparse Point Labels undergo refinement. We overlay original BBoxes, and within them, previously generated sparse point labels are replaced with labels corresponding to the BBox itself. This operation, termed Label Refinement by BBox, ensures the self-supervised learning model receives strong, focused, and class-specific guidance from the supervised model for targeted regions.

**STEP 4. kNN Label Propagation:** These automatically generated and refined Sparse Point Labels are then used as input for the self-supervised learning model. With DINOv2 features and HIL-Coral label propagation, more detailed segmentation masks are generated. In HIL-Coral label propagation, labels are propagated based on DINOv2-extracted feature vector $\mathbf{f_x} \in \mathbb{R}^D$ for each pixel x, using k-Nearest Neighbors (kNN). Cosine similarity between L2-normalized vectors is calculated.

$$S(\mathbf{f_a}, \mathbf{f_b}) = \frac{\mathbf{f_a} \cdot \mathbf{f_b}}{\|\mathbf{f_a}\|_2 \|\mathbf{f_b}\|_2} \quad (2)$$

The cosine similarity is effective for angular distance between feature vectors, making it robust to variations in magnitude and suitable for high-dimensional data. For each unlabeled pixel, we select k label-assigned pixels with highest similarity, defining kNN(x). We use k=3 (see Section 4.2). This k=3 choice enhances robustness against uncertainties in initial labels from USIS-SAM, improving upon k=1 suggested in the original HIL-Coral paper.

$$\text{kNN}(x) = \{x_{i1}, x_{i2}, \dots, x_{ik}\} \quad (3)$$

Based on kNN(x) majority vote, class $\hat{y}$ of x is determined. Here, $\mathbb{I}$ is an indicator function, returning 1 if belonging to class c and 0 otherwise.

$$\hat{y} = \underset{c \in \{C_1, C_2, \dots, C_N\}}{\text{argmax}} \sum_{i \in \text{kNN}(x)} \mathbb{I}(y_i = c) \quad (4)$$

This majority voting scheme aggregates local evidence from similar pixels to assign a consistent class label, enhancing robustness against noisy individual assignments. This automatic generation mechanism eliminates manual input in the Sparse

Points process of self-supervised learning, enabling full automation of seaweed segmentation.

### 3.3.2 MaskFusion

The MaskFusion method further refines the mask generated by SSL.Prop. (FIGURE 5). It selects highly reliable instances (seaweed regions) based on the segmentation mask and BBox information from SSL.Prop., considering both models' characteristics. Specifically, the MaskFusion process executes the following steps to intelligently merge detailed self-supervised segmentation with supervised class information:

**STEP 1. Instance Mask Conversion:** First, the semantic mask from SSL.Prop. (HIL-Coral label propagation result) is converted into an instance mask via a multi-step process. This involves encoding pixel colors/class IDs and identifying connected components. Minor noisy regions (e.g., those with an area below $1 \times 10^{-4}$% of total image pixels) are excluded, ensuring robust instance identification.

**STEP 2. Identification of Prioritized Dominant Instances:** Next, candidate BBoxes are selected from initial supervised learning outputs (USIS-SAM) based on a confidence score s(B) (>$T_s$, e.g., 0.8) and an area A(B) (<$R_{small}$, e.g., 0.10), similar to criteria in Equation (4). This selection emphasizes challenging instances like small seaweed. For each filtered BBox, the corresponding region of interest (ROI) is extracted from SSL.Prop.'s instance mask. Within this ROI, the most dominant instance (with the largest pixel count) is identified.

**STEP 3. Priority Mask Application:** An initial output mask is created as a copy of the USIS-SAM semantic mask. The entire area of each identified dominant instance (from STEP 2) is then pasted onto this initial output mask. This selective overriding integrates detailed, refined segmentation from the self-supervised branch for prioritized instances into a final mask respecting supervised class information and sub-pixel accuracy.

This approach improves segmentation accuracy for small and complex-shaped seaweed, often overlooked by supervised learning, by prioritizing features captured by self-supervised learning.

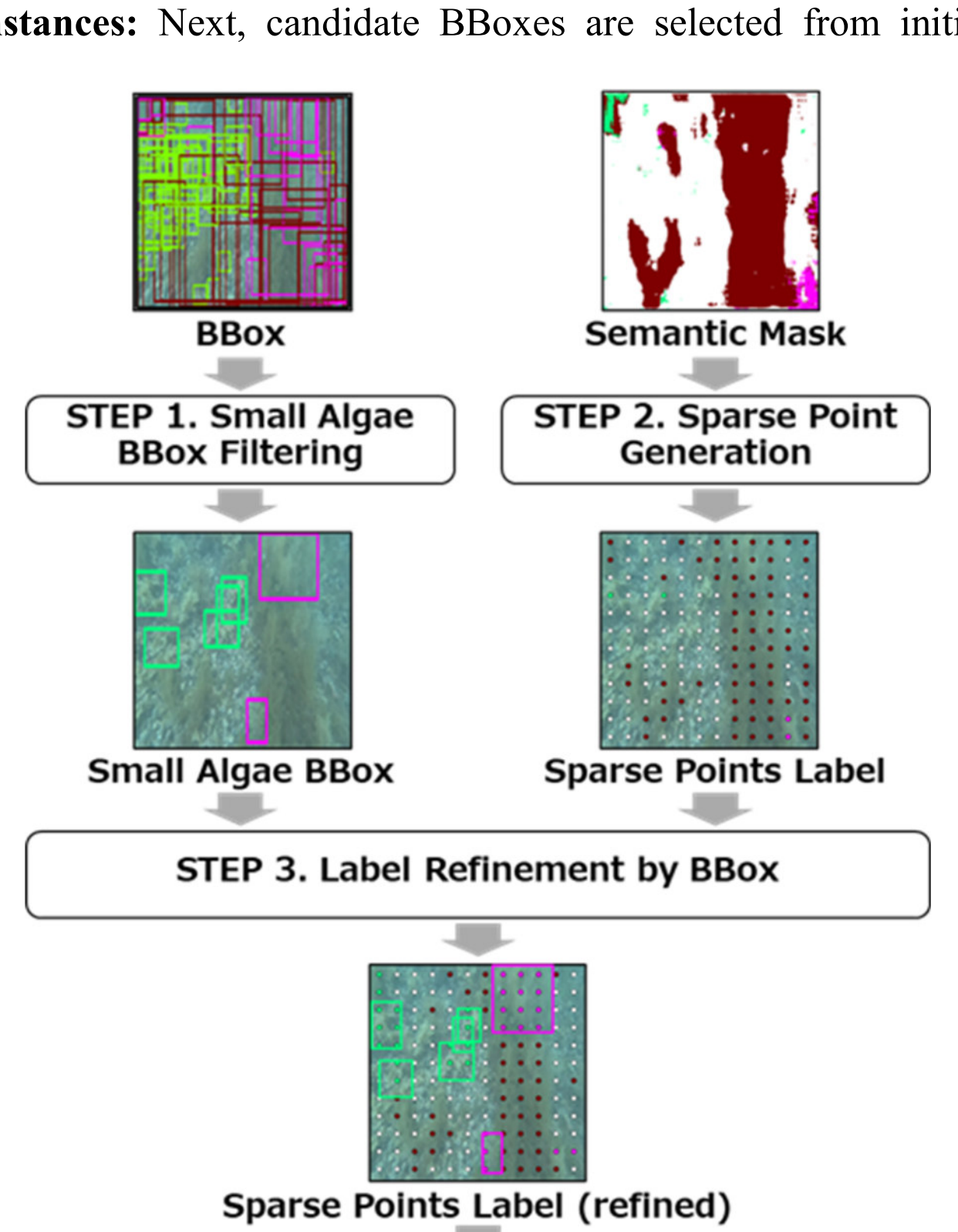

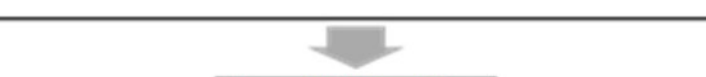

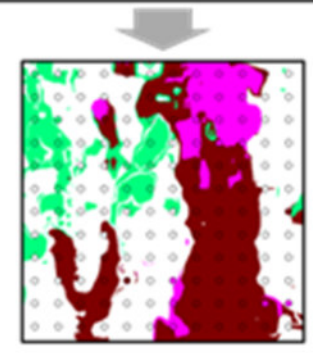

**FIGURE 4:** PROCESS OF SSL.PROP.

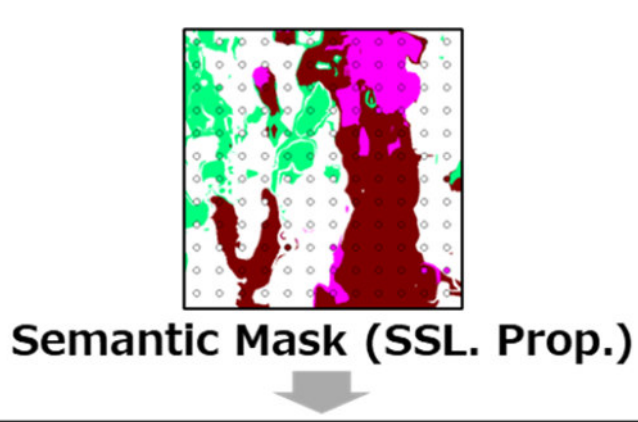

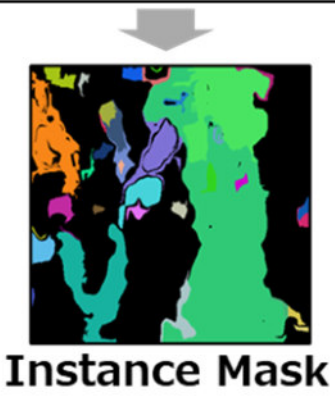

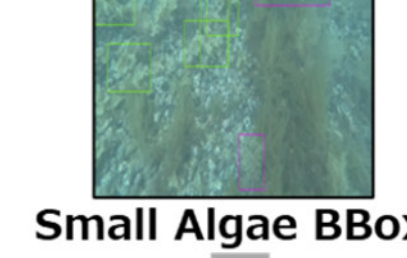

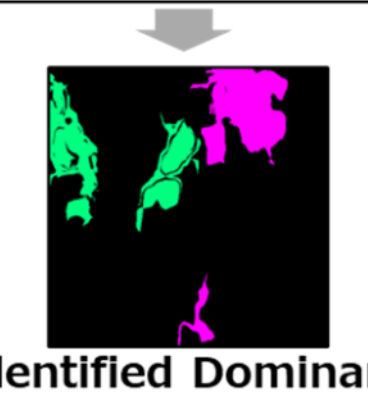

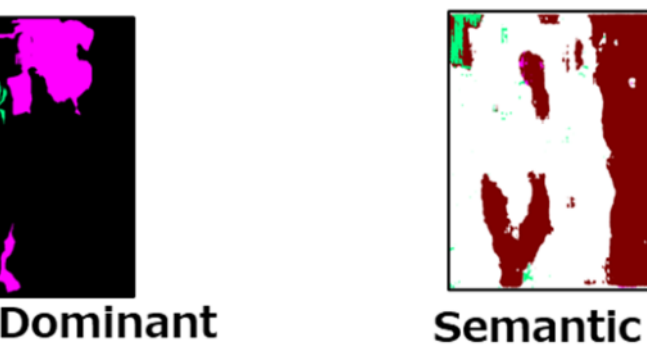

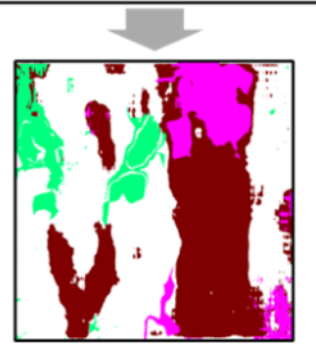

**FIGURE 5:** PROCESS OF MASKFUSION

## 4. EXPERIMENTS

This chapter describes the details of experiments conducted to verify the effectiveness of the proposed method. We will discuss dataset construction, implementation environment, and evaluation metrics.

### 4.1 Dataset

The dataset used in this research was constructed following the procedure below. Underwater images taken at a seaweed bed creation site in coastal areas of Yamaguchi Prefecture were used.

- **Shooting location and equipment:** Data collection was conducted at an artificial seaweed bed creation site off Kaminoseki, Yamaguchi Prefecture (FIGURE 6). Underwater camera GoPro 11 HERO Black mounted on an Autonomous Underwater Vehicle (AUV) Kyubic [7] was used for shooting. The AUV maintained an altitude of 1.5 [m], and the camera was set at a pitch angle of 45 [degrees]. Underwater videos were acquired while traversing four 15 [m] transects at 2 [m] intervals (FIGURE 7).
- **Data construction:** From the collected underwater videos, frames were selected at 100-frame intervals along the four transects of the AUV's trajectory, and 110 representative still images were extracted to construct the dataset. These images were divided into training (88 images), validation (12 images), and test sets (10 images) for deep learning model training, performance validation, and final evaluation, respectively. Each image was annotated with pixel-level high-precision annotations by experts and experienced annotators.
- **Target species:** The seaweed species targeted in this research are four representative classes that are highly relevant to blue carbon in coastal areas of Japan. The breakdown is as follows: *Sporochnus keyari*[8], *Sargassum piluliferum*, *Padina arborescens*, Background (all others)
- **Test data composition:** The pixel ratio of each class in the evaluation Test Set was: *Sporochnus keyari*: 38.7%, *Sargassum piluliferum*: 7.3%, *Padina arborescens*: 4.1%, Background: 49.9%. Based on the distribution characteristics of seaweed, *Sporochnus keyari* was defined as large seaweed (hereafter referred to as Large), and *Sargassum piluliferum* (hereafter referred to as SmallA) and *Padina arborescens* (hereafter referred to as SmallB) were defined as small seaweed, and their performance was evaluated separately.

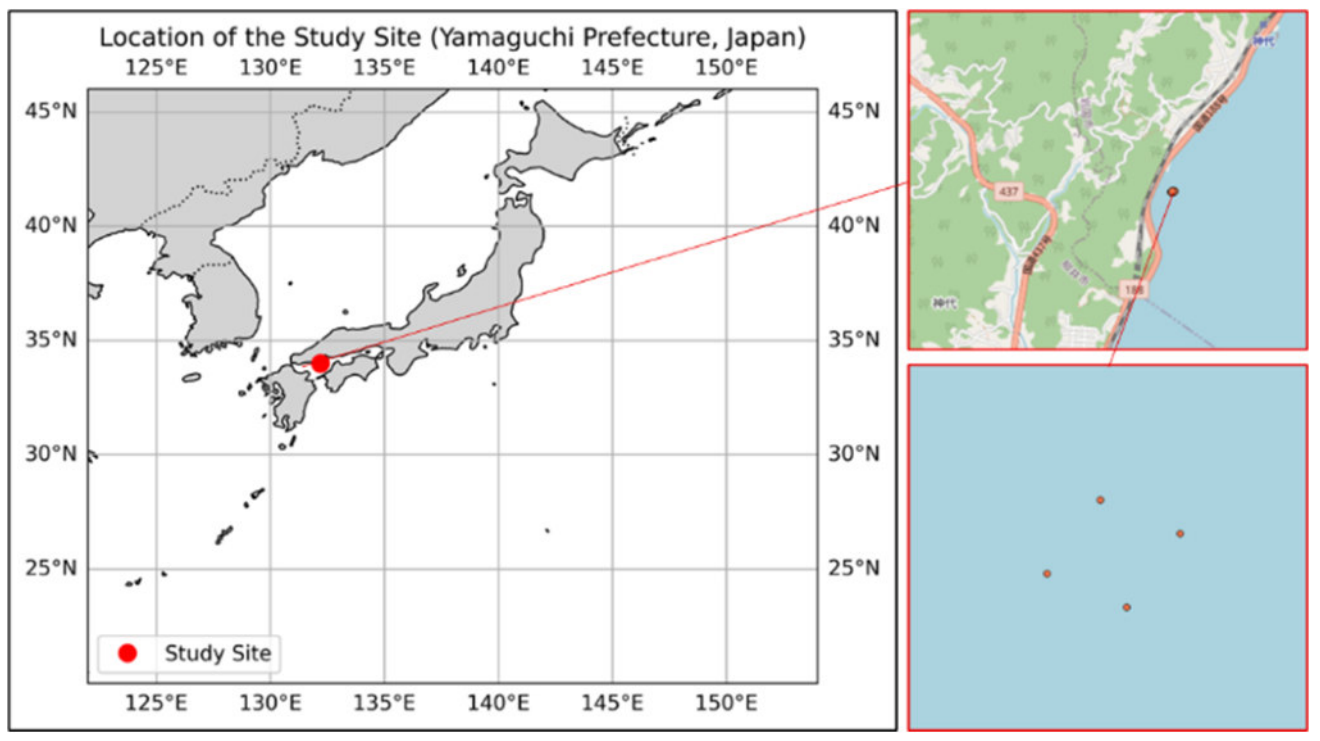

**FIGURE 6:** LOCATION OF STUDY SITE

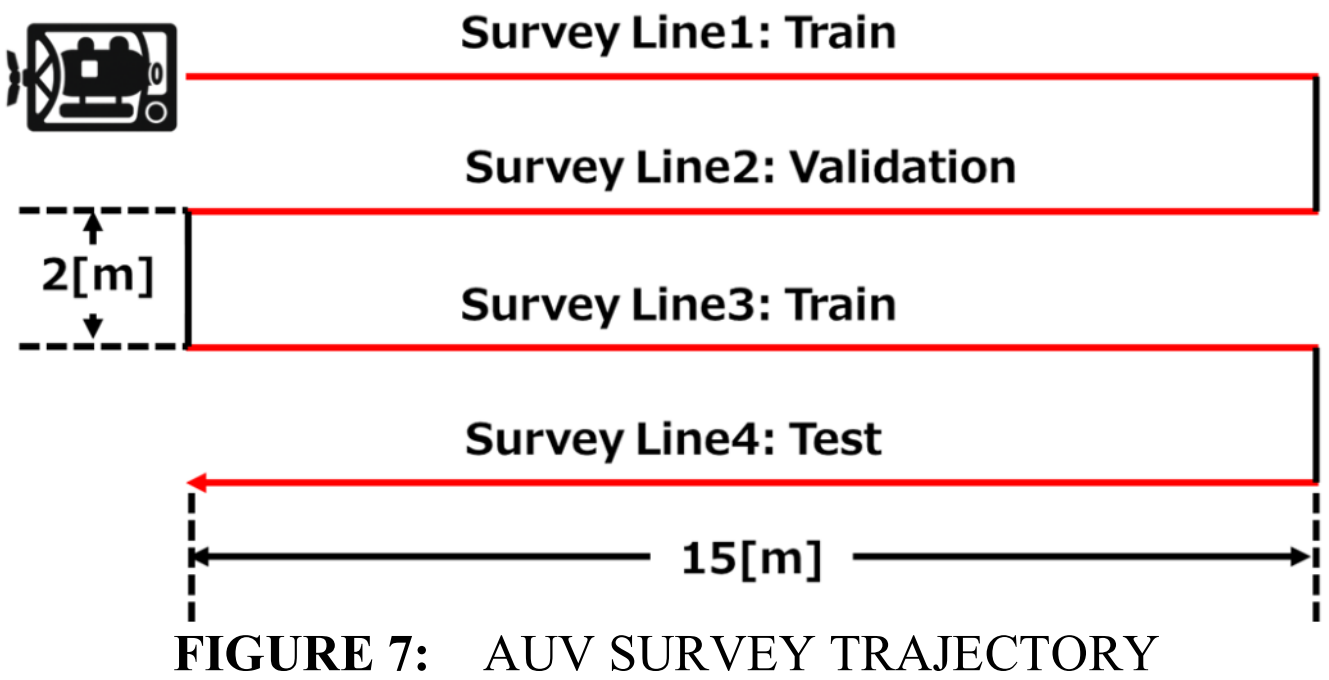

**FIGURE 7:** AUV SURVEY TRAJECTORY

### 4.2 Implementation

- **Development and Hardware Environment:** Development was carried out using Python 3.10.18 and PyTorch 2.1.2. All experiments were performed on a workstation equipped with NVIDIA A100 GPUs.
- **Supervised Learning Model:** The USIS-SAM model was adopted. USIS-SAM uses the ViT (Vision Transformer) backbone of SAM (Segment Anything Model) [9] pre-trained on the SA-1B dataset [9] and was fine-tuned with the training and validation datasets described in Section 4.1. The training conditions were: Optimizer: AdamW, Initial learning rate = 0.0001, Epochs = 24, Batch size = 2.
- **Self-Supervised Learning Model:** The DINOv2 model was adopted. DINOv2 is a large-scale web dataset called LVD-142M [4], curated by Meta, and was used without fine-tuning after pre-training by self-supervised learning. HIL-Coral was used for label propagation. HIL-Coral parameters were set to k=3 for k-Nearest Neighbors (kNN). This choice of k=3 was made to enhance robustness against uncertainties in initial labels derived from USIS-SAM outputs, departing from the k=1 suggested in the original HIL-Coral paper. The Sparse Points Label, which serves as input to self-supervised learning, was arranged in a uniform 12x12 grid within the image region, totaling 144 points.
- **Fusion Algorithm:** As common parameters for the proposed methods (SSL.Prop. and MaskFusion), a confidence score of 0.8 or higher for BBoxes output by the supervised learning model was used, and BBoxes whose area was 10% or less of the total image were judged as small seaweed. These thresholds were tuned through iterative experimentation on the validation set, aiming to optimally weigh the contributions of the supervised and self-supervised components. A confidence score of 0.8 ensures that only reliable detections from the supervised model initiate the sparse point generation, while the 10% area threshold for small seaweed helps prioritize the more granular segmentation capabilities of the self-supervised model for challenging smaller objects.

### 4.3 Comparative Methods

To evaluate the effectiveness of the proposed method, comparative experiments were conducted using three methods: USIS-SAM, SSL.Prop., and SSL.Prop.+MF(MaskFusion). USIS-SAM is a conventional method that serves as a baseline approach, utilizing only a supervised learning model to represent general seaweed segmentation performance. SSL.Prop., our initial proposed method, is a fusion method that automatically generates Sparse Points Labels from BBoxes and masks, directly adopting the self-supervised learning model's mask to integrate the benefits of automation and self-supervised learning. MaskFusion, our final proposed method, further refines the results of SSL.Prop. by selecting more important regions and pasting them onto the supervised learning model's masks, thereby maximizing the strengths of both models.

### 4.4 Evaluation Metrics

The performance of the proposed methods was evaluated using the following metrics, considering each target class independently. First, the basic terms used to define the evaluation metrics are shown below. These indicate the relationship between predicted results and ground truth labels in classification and segmentation tasks for individual classes.

- TP(True Positive): The number of pixels belonging to a specific class that are correctly classified as that class.
- TN(True Negative): The number of pixels not belonging to a specific class that are correctly classified as not belonging to that class.
- FP(False Positive): The number of pixels not belonging to a specific class that are incorrectly classified as belonging to that class.
- FN(False Negative): The number of pixels belonging to a specific class that are incorrectly classified as not belonging to that class.

Based on these definitions, the following evaluation metrics were calculated.

- **PA (Pixel Accuracy):** The percentage of all pixels that are correctly classified. This is the most basic metric showing overall classification performance.

$$PA = \frac{\sum_{i=1}^{C} TP_i}{\sum_{i=1}^{C} (TP_i + FN_i)} \quad (5)$$

- **mPA (mean Pixel Accuracy):** The average pixel-wise accuracy for each class. This metric is less affected by class imbalance.

$$mPA = \frac{1}{C} \sum_{i=1}^{C} \frac{TP_i}{(TP_i + FN_i)} \quad (6)$$

- **mIoU (mean Intersection over Union):** The average overlap (similarity) between the predicted region and the ground truth region for each class. This is one of the most widely adopted metrics for measuring segmentation accuracy.

$$mIoU = \frac{1}{C} \sum_{i=1}^{C} \frac{TP_i}{(TP_i + FP_i + FN_i)} \quad (7)$$

- **IoU_i (Intersection over Union for each class):** This is used to analyze the performance for specific seaweed species in detail.

$$IoU_i = \frac{TP_i}{TP_i + FP_i + FN_i} \quad (8)$$

- **Time:** The average processing time (in seconds) required to process one image. This indicates the automation capability and practical efficiency of the system.

## 5. RESULTS AND DISCUSSION

This chapter describes the results and discussion of the experiments conducted.

### 5.1 Results

Table 1 presents the experimental results, and FIGURE 8 shows the processing results. The remarkable effectiveness of our approach was confirmed. First, SSL.Prop. showed an improvement of +0.110 in mPA (0.532→0.642) and +0.065 in mIoU (0.449→0.514) compared to USIS-SAM. However, PA decreased by -0.012 (0.810→0.798). This point will be discussed in detail in the later section. Visually, FIGURE 8 illustrates these improvements by comparing the segmentation masks generated by each method across various underwater images. We can observe that both SSL.Prop. and SSL.Prop.+MF produce more complete and accurate segmentation boundaries for seaweed, especially for smaller or irregularly shaped instances, compared to the baseline USIS-SAM. Next, SSL.Prop.+MF further improved accuracy from SSL.Prop., achieving an additional improvement of +0.017 in mIoU (0.514→0.531). As a result, SSL.Prop.+MF outperformed USIS-SAM in all key metrics: PA (0.818), mPA (0.644), and mIoU (0.531). Regarding processing time, SSL.Prop.+MF showed a significant increase compared to USIS-SAM: SSL.Prop. increased by +4.18 seconds (0.865 seconds→5.041 seconds), and SSL.Prop.+MF further increased by +0.24 seconds (5.041 seconds→5.283 seconds). This increase in processing time reflects the increased computational cost due to the two-stage processing and fusion algorithm. However, considering the significant gains in segmentation accuracy, particularly for challenging fine-grained tasks in underwater environments, this increase in processing time is deemed acceptable within the context of most blue carbon assessment applications where strict real-time performance might not be the primary requirement.

Furthermore, Table 2 shows the comparison of IoU results for small seaweed such as Large, SmallA and SmallB. From Table 2, USIS-SAM showed a relatively high IoU for Large (0.701) but significantly lower IoUs for SmallA (0.175) and SmallB (0.146). This highlights a major limitation of conventional supervised approaches when dealing with fine-grained classification problems or rare species within the training data. In contrast, SSL.Prop. showed a slight decrease in IoU for Large (-0.009, 0.701 → 0.692) but significant improvements for SmallA (+0.147, 0.175→0.322) and SmallB (+0.161, 0.146→0.307). These substantial gains in SmallA and SmallB indicate that the self-supervised learning components, by effectively leveraging image internal features, are highly successful in identifying and segmenting smaller and morphologically complex seaweed species. Finally, SSL.Prop.+MF generally maintained the effectiveness for small seaweed of SSL.Prop., with a slight decrease in SmallB but improvements in SmallA, while also improving IoU for Large (+0.022, 0.692 → 0.714). This collective improvement across different seaweed categories underscores SSL.Prop.+MF's ability to achieve a more balanced and robust segmentation performance, adapting to varying sizes and complexities of target objects by intelligently combining the strengths of both learning paradigms. This confirmed that the balance of accuracy across classes was improved.

**TABLE 1:** PERFORMANCE COMPARISON OF METHODS

| Method | PA | mPA | mIoU | Time per Image[s] |
|---|---|---|---|---|
| USIS-SAM(Base) | 0.810 | 0.532 | 0.449 | **0.865** |
| SSL.Prop.(Ours) | 0.798 | 0.642 | 0.514 | 5.041 |
| SSL.Prop.+MF(Ours) | **0.818** | **0.644** | **0.531** | 5.283 |

**TABLE 2:** IOU PERFORMANCE FOR DIFFERENT SEAWEED CATEGORIES

| Method | Large | SmallA | SmallB | Back ground |
|---|---|---|---|---|
| USIS-SAM(Base) | 0.701 | 0.175 | 0.146 | 0.775 |
| SSL.Prop.(Ours) | 0.692 | 0.322 | **0.307** | 0.736 |
| SSL.Prop.+MF(Ours) | **0.714** | **0.335** | 0.302 | **0.771** |

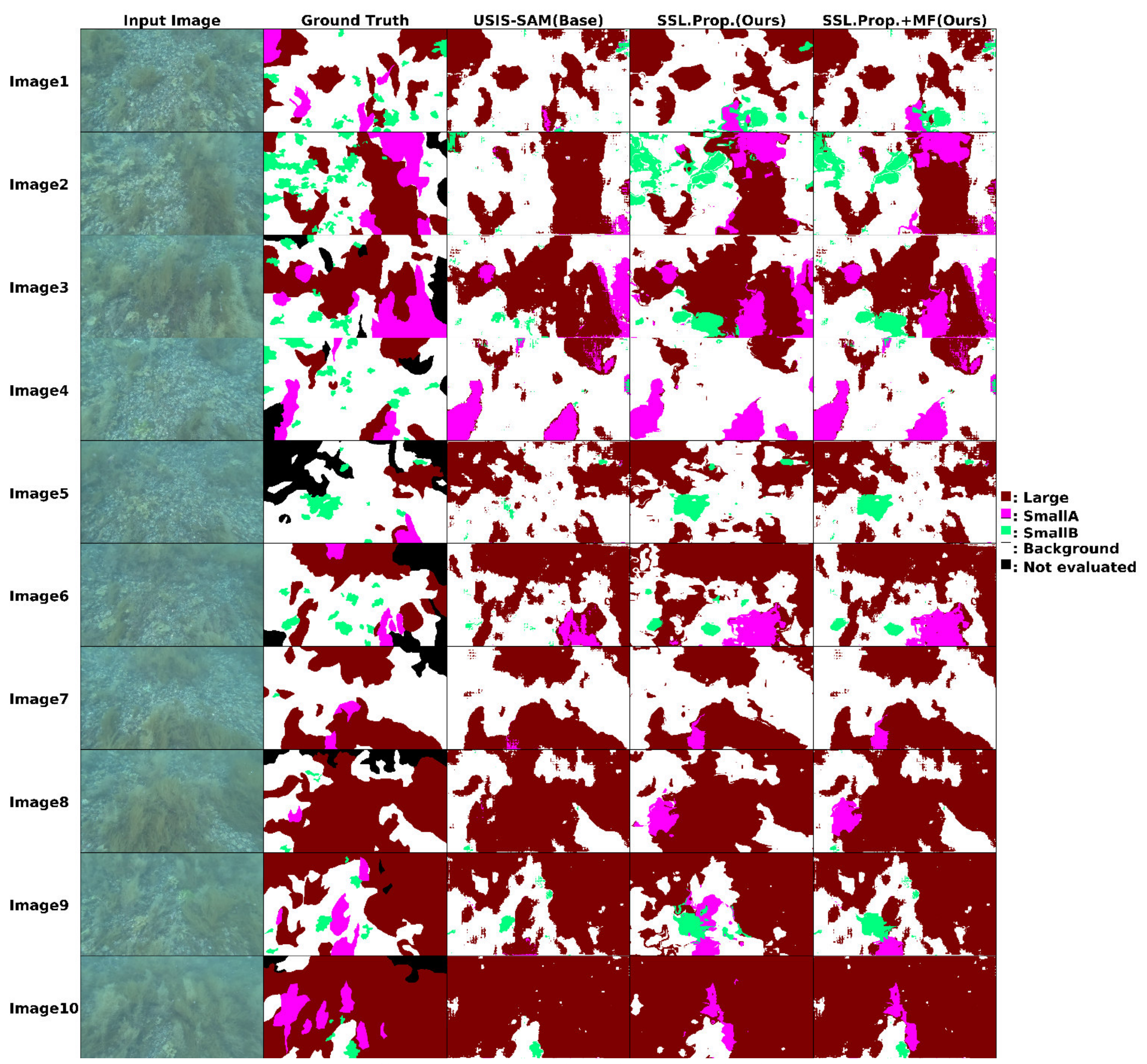

**FIGURE 8:** RESULTS OF SEMANTIC MASK

### 5.2 Discussion

This section discusses the findings from the experimental results and the remaining challenges. The performance analyses presented in Tables 1 and 2, combined with the visual evidence from FIGURE 8, provide a comprehensive understanding of the proposed SSL.Prop.+MF model's efficacy. Our observations consistently support the initial hypotheses regarding the complementary strengths of supervised and self-supervised learning for nuanced underwater image segmentation. This discussion aims to dissect these findings, offering insights into the underlying reasons for the performance gains and addressing the practical implications for real-world applications in marine ecosystem monitoring and blue carbon assessment.

**Challenges of Conventional Methods and Effectiveness of Proposed Methods:** The results from conventional methods (USIS-SAM) clearly showed a tendency for lower IoU in small seaweed species compared to large seaweed, which is attributable to the inherent weakness of supervised learning models—diminished performance for specific classes due to limited training data. In contrast, the proposed method (SSL.Prop.) improved the IoU for small seaweed. This result clearly demonstrates that the robustness of self-supervised learning against domain gaps and the automated process of Sparse Points Label generation from BBox information, complement the weaknesses of supervised learning and are highly effective for recognizing rare and small species. The slight decrease in PA is considered to be due to SSL.Prop.'s strong focus on improving the accuracy of small seaweed, resulting in a limited impact on pixels constituting the background and other features.

**Further Accuracy Improvement with SSL.Prop.+MF Method:** SSL.Prop.+MF method improved IoU for 3 out of 4 classes compared to SSL.Prop., suggesting that it appropriately selects more reliable and accurate instances by considering the characteristics of both models based on the masks and BBox information obtained from SSL.Prop., and integrates them into the final segmentation result. In particular, the improvement in IoU for large seaweed also supports that the SSL.Prop. + MF method captures the characteristics of individual seaweed instances in more detail and effectively utilizes the complementary relationships between the outputs of both models.

**Processing Time and Practicality:** While the processing time of the proposed SSL.Prop.+MF method is longer than that of conventional methods, in applications such as coverage surveys for blue carbon assessment, real-time processing is not always strictly required. Therefore, an increase in processing time of a few seconds is considered to be a sufficiently acceptable level, given the value and cost-effectiveness of the highly accurate segmentation results obtained. Furthermore, from a future perspective, it is expected that the generated highly accurate segmentation results from the SSL.Prop.+MF method can be utilized as new teaching data to retrain more efficient supervised deep learning models, thereby reducing processing time. This approach has the potential to reduce the effort required for manual annotation and support the construction of larger datasets.

**Remaining Challenges and Future Directions:** While the proposed SSL.Prop.+MF method demonstrated high performance, some remaining challenges like undetected and over-detected instances were identified.

- **Undetected Instances:** Some seaweed instances were undetected (e.g., SmallB in Image1, as shown in FIGURE 8). This can be attributed to two main causes: first, if the BBox of the seaweed was not detected by the initial supervised learning model (USIS-SAM) (FIGURE 9), and second, if the Sparse Points Label generation mechanism was insufficient, failing to capture subtle features or ambiguous contours within the seaweed image. As countermeasures for the former, adjusting the supervised learning model itself to be more sensitive or combining it with auxiliary detection methods can be considered. For the latter, it is possible to increase the number of Sparse Points Labels generated, but this directly leads to an increase in processing time. A more efficient approach would be to define Sparse Points in two stages: rough Sparse Points for the entire image, and detailed Sparse Points for within the BBox detected by supervised learning, thereby generating Sparse Points Labels in a more targeted manner, which could improve accuracy while suppressing computational costs.
- **Over-detected Instances:** In some cases, over-detection occurred where regions that were not target seaweed were incorrectly segmented (e.g., SmallA and B in Image9, as shown in FIGURE 8). This suggests that in label propagation from Sparse Points, simply applying the same label to all regions within the BBox may have included elements other than the target (FIGURE 10). As a solution, the processing within the BBox needs to be further refined. Specifically, applying general image processing techniques within the BBox (e.g., binarization, foreground/background separation algorithms) to extract only the true seaweed parts and applying labels only to those foreground regions could prevent unnecessary label propagation and suppress over-detection.

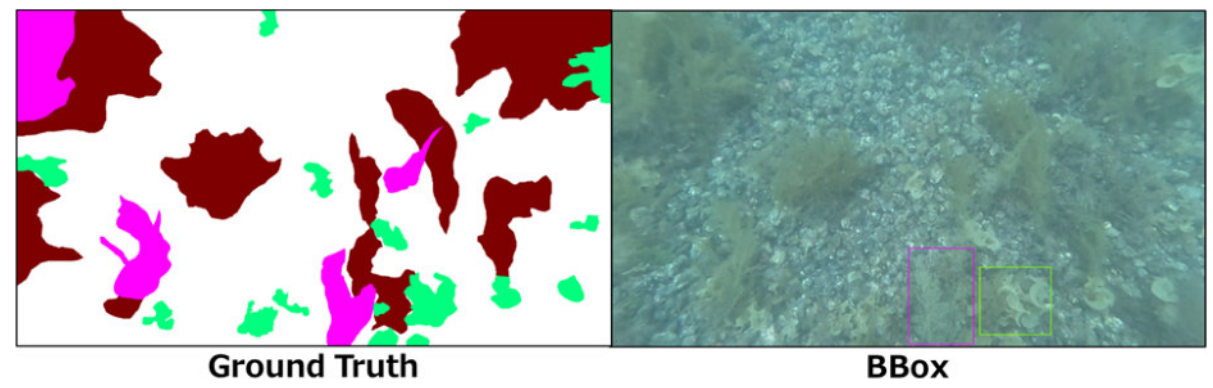

**FIGURE 9:** UNDETECTED INSTANCES

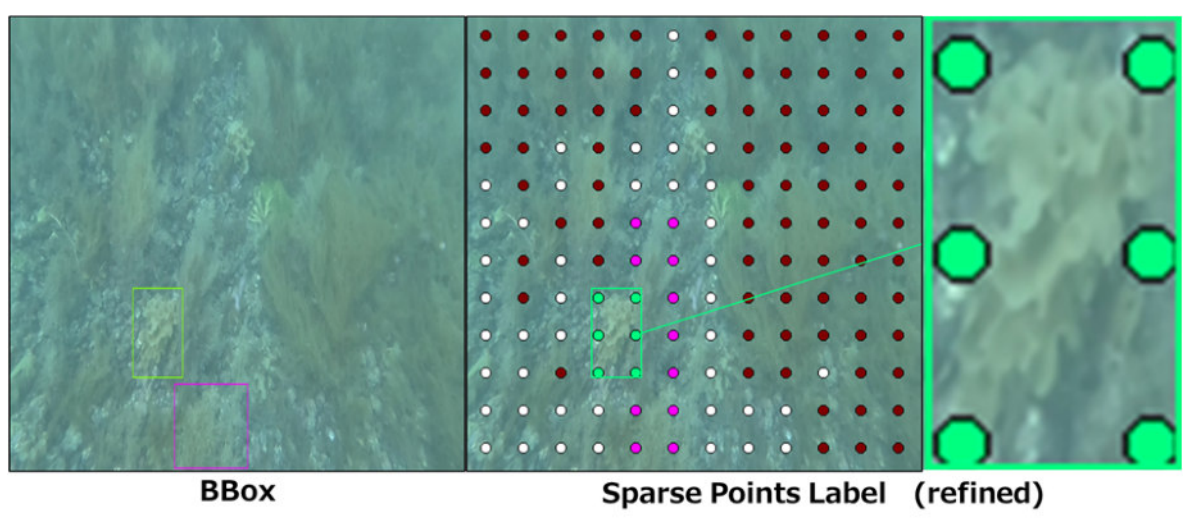

**FIGURE 10:** OVER-DETECTED INSTANCES

## 6. CONCLUSION

In this research, we proposed a novel two-stage seaweed segmentation technique that complementarily utilizes the strengths of supervised and self-supervised learning to improve the accuracy of blue carbon assessment for Ocean Digital Twin realization. Through multiple comparative experiments, we confirmed that our full proposed method (SSL.Prop.+MF) significantly improves segmentation accuracy compared to USIS-SAM, especially for targets including small seaweed species where training data is often insufficient, while also demonstrating the potential for process automation. Our full proposed method enables the automatic assignment of class labels, which was difficult to achieve with self-supervised learning, by leveraging class information obtained from supervised learning. It also reduces the impact of domain gaps by estimating the approximate location with supervised learning and then performing self-supervised learning within test images, achieving high-accuracy segmentation. Specifically, our full proposed method constructed a technique that estimates the approximate location of each target species as sparse point information through instance segmentation by supervised learning, and then automatically segments detailed regions for each target species using self-supervised learning based on those results. This technology is expected to improve the accuracy of quantitative blue carbon assessment and contribute to efficient, large-scale marine ecosystem monitoring. This will contribute to the effective utilization of blue carbon as a climate change countermeasure and to marine environmental conservation.

## ACKNOWLEDGEMENTS

This research benefited from the generous support and cooperation of many individuals. We would like to express our deepest gratitude here. Seiji Yamada and Yuji Ano of the Yamaguchi Prefectural Industrial Technology Institute provided invaluable guidance and support for the overall management and promotion of data collection in this research. Kazuo Ishii and Yuya Nishida of Kyushu Institute of Technology provided significant cooperation in the collection of underwater video data using AUVs. Furthermore, Noboru Murase, Mahiko Abe, and Kazuhiro Tokunaga of National Fisheries University contributed to the annotation work for the collected video data. We are also sincerely grateful to Kenji Sugimoto of National Institute of Technology, Ube College and Yasuhito Miyata, Kentaro Iwai of JFE Steel Corporation for kindly providing access to the artificial seaweed bed creation site, which served as our shooting location. We deeply appreciate the valuable contributions of all.